\documentclass{article}
\usepackage{kr}

\usepackage{times}
\usepackage{xcolor}
\usepackage{soul}
\usepackage{url}
\usepackage[hidelinks]{hyperref}
\usepackage[utf8]{inputenc}
\usepackage[small]{caption}
\usepackage{graphicx}
\usepackage{amsmath}
\usepackage{amsthm}
\usepackage{comment}
\usepackage{booktabs}
\usepackage{algorithm}
\usepackage{algorithmicx}
\usepackage{graphicx}
\usepackage[T1]{fontenc}
\usepackage{algpseudocode}
\usepackage{xspace}
\usepackage{textgreek}
\usepackage{tabularx}   
\usepackage{booktabs}    
\usepackage{pifont}     
\usepackage{xcolor}
\usepackage{xurl}
\newcommand{\vadalog}{\textsc{Vadalog}\xspace}

\newcommand{\vadaorchestra}{\textsc{VadaOrchestra}\xspace}

\usepackage{enumitem}

\title{VADAOrchestra: Neurosymbolic Orchestration of Adaptive Reasoning Workflows} 

\author{%
Teodoro Baldazzi$^1$\and
Luigi Bellomarini$^2$\and
Andrea Coletta$^2$\and
Michela Iezzi$^2$\\
Carsten Maple$^3$\and
Alessandro Pesare$^1$\and
Emanuel Sallinger$^1$ \\
\affiliations
$^1$TU Wien\\
$^2$Banca d'Italia\\
$^3$University of Warwick\\
\emails
\{teodoro.baldazzi, alessandro.pesare, emanuel.sallinger\}@tuwien.ac.at,
\{luigi.bellomarini, andrea.coletta, michela.iezzi\}@bancaditalia.it,
CM@warwick.ac.uk
}
\begin{document}

\maketitle

\begin{abstract}
Decision-making in real-world settings rarely follows a fixed script. Instead, it unfolds as a dynamic reasoning process in which the appropriate course of action evolves as new context and data become available. Traditional Business Process Management systems provide rigor, determinism, and auditability, yet they generally struggle to adapt their execution at runtime. Conversely, agentic systems based on Large Language Models (LLMs) bring flexibility to decision-making, but they are inherently opaque, often unreliable, and suffer from significant scalability constraints when operating over large datasets.
To combine these complementary paradigms, we introduce VADAOrchestra, a neurosymbolic framework that models complex workflows as evolving reasoning processes. The framework adopts a hybrid approach: given a user query and a collection of data sources, an LLM-based orchestrator incrementally plans and adapts the workflow. This is encoded as a logic program in a fragment of Datalog+/- where predicates correspond to tool invocations and rules represent both predefined domain dependencies and logic constructs synthesized on demand to manipulate intermediate results. All logical inference tasks are then executed by a state-of-the-art Datalog+/- symbolic engine. This approach provides a verifiable reasoning trace, supporting the auditability and reproducibility of the entire process. Furthermore, by decoupling high-level orchestration from symbolic inference, it addresses scalability concerns, enabling complex reasoning over large datasets through targeted data querying. We evaluate VADAOrchestra on real-world financial use cases, demonstrating faithfulness, scalability, and explainability compared to standard agentic architectures.
\end{abstract}

\section{Introduction}
\label{sec:introduction}
Real-world decision-making processes, particularly in complex domains such as finance, cannot be reduced to static, predefined workflows. 
In these knowledge-intensive settings, a workflow is not merely a sequence of tasks, but a dynamic process in which the appropriate course of action depends on newly available information, contextual factors, and intermediate outcomes.
The central challenge in automating such workflows arises from their dual nature: they are partially structured, following well-defined procedures derived from domain expertise, and partially dynamic, branching unpredictably based on runtime findings. 

Historically, Business Process Management (BPM) systems have provided the essential rigor, transparency, and compliance guarantees required for highly structured enterprise workflows. However, their largely static nature proves to be a major limitation, as these frameworks struggle to accommodate the data-driven branching that characterizes dynamic workflows~\cite{diciccio2015knowledge}. 

More recently, agentic AI has emerged as a foundational paradigm for automating knowledge-intensive workflows~\cite{wei2026agenticreasoninglargelanguage}. Leveraging Large Language Models (LLMs), agentic systems exhibit strong capabilities in context-aware decision-making, planning, and tool orchestration~\cite{10.5555/3666122.3666499,10.5555/3737916.3741936}.
LLMs can interpret high-level goals, decompose them into sub-tasks, and interact with external resources such as web search engines, databases, and software tools. 
Standardized protocols, such as the Model Context Protocol (MCP), further facilitate this interaction by exposing tools through a uniform interface, enabling LLMs to select and compose tool invocations dynamically at runtime~\cite{anthropic2024mcp}.

Despite their capabilities, LLM-based systems still face a fundamental limitation in explainability. Research on faithfulness has shown that LLMs frequently generate post-hoc explanations that do not reflect their actual reasoning paths~\cite{Lanham2023MeasuringFI,turpin2023language,Tanneru2024OnTH,DBLP:conf/iclr/MattonNGK25}.

\begin{figure*}[t]
    \centering
    \includegraphics[width=1\textwidth]{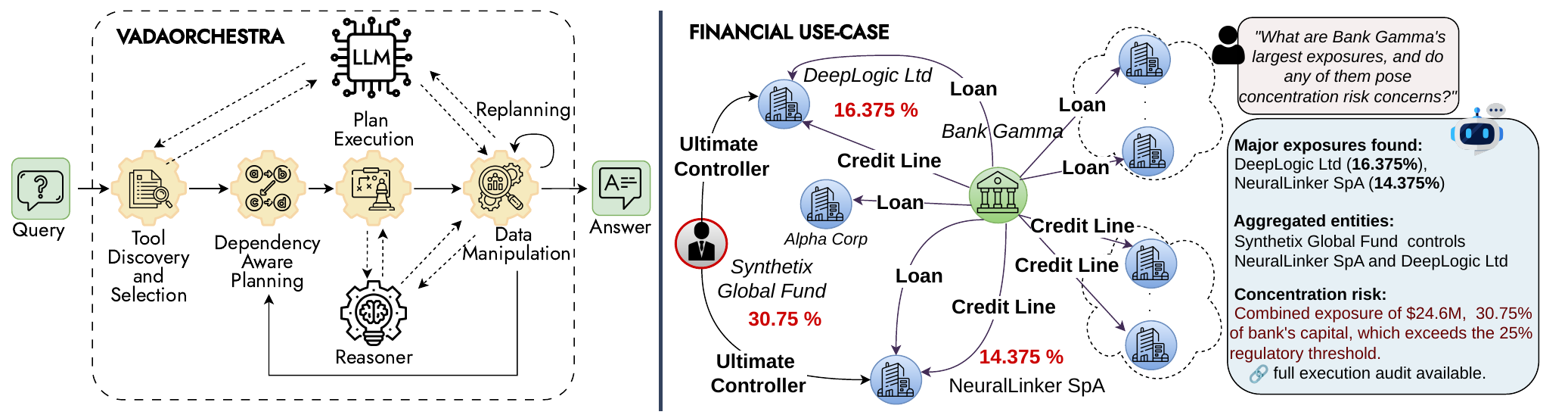}
    \caption{The \vadaorchestra framework, with the system architecture (left), and a real financial use case (right). The example shows a concentration-risk assessment for Bank Gamma under a 25\% capital threshold. The framework processes thousands of exposures and identifies two major exposures, namely \textit{DeepLogic Ltd} and \textit{NeuralLinker SpA}. Then, the framework performs a group-level assessment, and through structured data manipulation and reasoning tools,  correctly determining that the ultimate controller, \textit{Synthetix Global Fund}, breaches the regulatory threshold. The output includes a fully auditable trace, as the entire workflow is represented as a logical reasoning process.
    }\label{fig:runningexample}
\end{figure*}

\smallskip \noindent
\textbf{A Central Bank dynamic workflow.}
As an illustrative case, consider the following scenario from a central bank in Europe, regarding the assessment of a bank's ``concentration risk''. This activity requires the analyst to identify large exposures (i.e., significant amounts of money lent or committed to individual counterparties) and determine whether any exceed regulatory limits, thereby preventing excessive risk from a single entity. As illustrated in Figure~\ref{fig:runningexample}, even a relatively contained case involves multiple counterparties, heterogeneous credit instruments (loans and credit lines), and layered ownership structures. Indeed, realistic deployments scale this complexity to thousands of entities. Rather than a static compliance check, this assessment is a dynamic investigation. While the components of an exposure, such as loans and credit lines, are defined by domain knowledge, the decision-making process must unfold adaptively based on runtime findings. If a single counterparty directly exceeds the regulatory threshold, a breach is identified and the analysis can terminate immediately. However, if no individual breach is detected, the reasoning shifts toward a more complex investigation. The challenge is not merely arithmetic. As shown in Figure~\ref{fig:runningexample}, \textit{DeepLogic Ltd} (16.375\%) and \textit{NeuralLinker SpA} (14.375\%) appear as fully independent counterparties, each individually compliant with the 25\% regulatory limit. There is no a priori signal that they should be considered together. Only by actively traversing their ownership chains does it emerge that both are ultimately controlled by \textit{Synthetix Global Fund}, at which point aggregation reveals a combined exposure of 30.75\%, constituting a clear breach. This ownership structure cannot be known in advance: it emerges only at runtime, as the investigation traverses corporate linkages among seemingly independent counterparties. The objective is to determine whether multiple individually compliant exposures actually form an aggregated risk group and whether their interconnected risks collectively breach the regulatory threshold once aggregated. 

\noindent
\textbf{Limits of current approaches.}
Even if an LLM-based agent correctly identifies that a specific counterparty poses a concentration risk, the reasoning behind this decision would remain opaque. \textit{Which logic underlies the assessment? Were all types of credit considered appropriately? Was the potential counterparty’s affiliation with a larger corporate group taken into account?} Without an auditable trace, the analyst cannot verify the reasoning process or assess whether a correct conclusion resulted from the right reasons. This opacity fundamentally undermines trust in high-stakes domains. 
\\
Beyond explainability, scalability poses equally relevant issues. LLM context windows are finite, limiting the amount of data that can be processed. Empirical studies show that, as context length increases, models exhibit degraded attention to information in the middle of the context (the so-called ``\textit{lost in the middle}'' phenomenon~\cite{liu-etal-2024-lost}), leading to overlooked details and compromised decisions. 
\\
Even recent multi-agent paradigms, such as chain of agents~\cite{zhang2024chain} and agent loops~\cite{wu2024autogen}, where multiple specialized agents collaborate or iterate to handle long-horizon tasks, do not fully resolve all these issues. Indeed, individual agents still operate within finite context windows, inter-agent communication introduces coordination overhead and information loss, and the reasoning process remains distributed across opaque generative steps.
\\
In our concentration risk example, a realistic analysis may involve thousands of exposure records across multiple counterparties, each possibly requiring an investigation of corporate group structures. Naively loading all the data into the context window is certainly infeasible. At the same time, defining retrieval criteria in advance is non-trivial, as overly restrictive filters risk omitting critical information.
\\
\noindent
\textbf{A hybrid solution.}
To effectively address these challenges, we propose \vadaorchestra, a neurosymbolic framework that leverages an LLM orchestrator for explicit, auditable, and dynamically adaptable decision-making workflows. Given a complex user query (e.g., an assessment of a bank’s concentration risk) and a set of data sources, the system is required not only to produce an answer, but also to expose the verifiable reasoning steps that justify it. 
To this end, we model the workflow itself as a logical reasoning process. Specifically, we represent the entire process, including both structured and dynamic components, as an evolving logic program in Vadalog~\cite{bellomarini2018vadalog}, a Datalog$^{\pm}$~\cite{cali2012general} -based declarative language.
In this program, predicates correspond to MCP tool invocations, rules encode dependencies among them, and inputs and outputs of the rules are the bindings of the predicates.
In this way, every step of the workflow is captured in a logical trace, enabling full auditability and reproducibility of the reasoning process.
\\
The structured component of the process is guided by a \textit{dependency graph} that encodes the logical relationships between domain concepts, providing the deterministic backbone required to orchestrate the reasoning process and enforce formal constraints.
Once a high-level concept (e.g., bank's exposures) is identified from the user’s query, the system consults the dependency graph to define an initial execution plan. This ensures the execution order of MCP tools is enforced, rather than leaving it up to the LLM to infer.
\\
The dynamic component is managed by an LLM-based \textit{orchestrator} that adapts the workflow in response to intermediate results. The reasoning flow is data-driven: the values of derived facts determine which tools are invoked next, their parameters, or whether the process should terminate. Furthermore, the orchestrator can synthesize Vadalog rules on the fly to perform data manipulations, such as joins and aggregations, that only become relevant at runtime.

\smallskip \noindent
\textbf{Benefits of the proposal.}
This hybrid approach provides three main benefits by blending the operational rigor of BPM systems with the adaptivity of Agentic AI. 
First, by rooting the workflow in a dependency graph, the system strictly adheres to domain-specific constraints. Unlike purely agentic approaches that may ``hallucinate'', our framework enforces the deterministic behavior typical of logic-based systems, ensuring that high-stakes analyses (e.g., concentration risk) follow well-defined procedures. At the same time, adaptivity is preserved: the LLM-based orchestrator manages planning complexity by invoking tools, adapting the reasoning flow, and synthesizing logic operations on demand to handle contingencies that cannot be fully predefined.

Second, representing the entire process, both the structured and dynamic components, as a Vadalog logic program provides full transparency. The resulting logical trace makes every reasoning step explicit, enabling auditability and allowing analysts to verify not only the final outcome, but also the correctness of the reasoning that led to it. Moreover, by decoupling the orchestration and planning, handled by the LLM, from the actual data inference, executed by dedicated MCP tools, the framework ensures that the results are also reliable and reproducible for the analysts.

Finally, this architecture addresses typical scalability limitations of traditional agentic systems, where performance often degrades as the volume of information increases. The orchestrator inspects the cardinality and statistical distribution of the intermediate results to formulate fine-grained data querying strategies. Based on these high-level summaries, it selectively retrieves only the data subsets required for the next stage of planning.

\smallskip \noindent
\textbf{Contributions.}
In this paper, we present \vadaorchestra, a neurosymbolic framework that bridges BPM rigor with the adaptability of Agentic AI. The main contributions of this work are as follows: 
\begin{itemize} 
    \item We design \textbf{a hybrid architecture that integrates logic-based reasoning with dynamic LLM-driven orchestration} to automate knowledge-intensive workflows.
    \item We propose \textbf{a methodology for the on-demand synthesis of process rules in Vadalog}, enabling data manipulation over intermediate results that emerge at runtime.
    \item We experimentally showcase the effectiveness of our approach on \textbf{a real-world financial use case}, demonstrating its scalability and the faithfulness of its reasoning traces over standard agentic architectures.
\end{itemize}

\section{System Overview}
\label{sec:overview}
Knowledge-intensive workflows in domains such as finance require both adherence to structured procedures and the flexibility to adapt to intermediate findings. Figure~\ref{fig:runningexample} illustrates our framework, \vadaorchestra. It automates such workflows through an LLM-based orchestrator for dynamic decision planning and a symbolic reasoning engine for logical inference, providing adaptive yet auditable executions.

Architecturally, the framework enforces a clear separation between planning and execution, thanks to a tight interaction between the LLM-orchestrator and a symbolic reasoner via MCP tools. The orchestrator is equipped with a set of tools, including reasoning tools (e.g., computing exposures) and support tools (e.g., navigating and manipulating intermediate data) whose composition is guided by a dependency graph to retrieve information and answer user queries. A reasoning tool is defined as a logic program in Vadalog, that is, a set of \textit{facts} and \textit{rules} encoding domain-specific inference logic, with support for aggregations, arithmetic expressions, and negation. Further details regarding the Vadalog language are discussed in Section~\ref{sec:preliminaries}.

\subsection{A Financial Use Case}
We consider the scenario depicted in Figure~\ref{fig:runningexample}, where the analyst of a Central Bank in Europe assesses whether a commercial bank complies with concentration risk limits. In banking regulation, an \textit{exposure} represents the total credit risk a bank holds toward a counterparty, including loans, credit lines, and other commitments, which may pose significant risk when exceeding a regulatory threshold. In our example, the analyst may ask the following question:
\begin{quote}
\textit{“What are Bank Gamma’s largest exposures, and do any of them raise concentration risk concerns?”}
\end{quote}
This question highlights the adaptability of our framework to a complex workflow, which consists of an initial \textit{structured phase} — where exposures are computed according to predefined domain rules — and a subsequent \textit{dynamic phase} driven by the actual findings. 

\smallskip \noindent
\textbf{Structured Phase.}
The workflow starts when the LLM-based orchestrator receives the analyst’s question. It first discovers the available MCP tools and performs semantic matching between the query and the tool specifications to identify the most appropriate ones to invoke. In this case, the question mentions ``exposures'' and ``concentration risk concerns'', thus the LLM-orchestrator selects \textit{get\_bank\_exposures} and \textit{get\_bank\_info}, extracting the parameters from the query (e.g., \textit{``Bank Gamma''}) to retrieve the bank’s capital and regulatory thresholds. 

Before executing any tools, the system invokes the support tool \textit{get\_dependencies} to identify inter-tool dependencies. High-level reasoning tools can in fact be expressed in terms of reusable lower-level tools, which must be executed in a meaningful order. For instance, \textit{get\_bank\_exposures(bank\_name)} depends on \textit{get\_loans(bank\_name)} and \textit{get\_credit\_lines(bank\_name)}, which provide the data required to compute the exposures. This step enables the orchestrator to consolidate the execution plan (i.e., first block of Figure~\ref{fig:trace}), which is then executed.
For each execution of a \textit{reasoning tool}, the \vadalog engine~\cite{2022VadaArchitecture,bellomarini2024vadalog} executes the associated program and materializes the results in a persistent storage, allowing the LLM to navigate and manipulate them via support tools (see Step 4 of Figure~\ref{fig:trace}) without saturating the context window.
\\
\textbf{\textit{The structured phase returns two major exposures:}} \textit{NeuralLinker SpA} (11.5M) and \textit{DeepLogic Ltd} (13.1M), corresponding to 14.375\% and 16.375\% of Bank Gamma’s capital, respectively. Both values are below the regulatory threshold of 25\%. A traditional static BPM system would typically terminate at this point and report a \textit{compliant} status together with the computed exposures.  

\smallskip
\noindent\textbf{Dynamic Phase.}
In our framework, after each execution plan is carried out, the LLM orchestrator analyzes the actual findings to determine whether refinements are necessary, in an iterative process referred to as the \textit{replanning} stage. 

In the running example, the orchestrator detects a semantic similarity between the profiles of \textit{NeuralLinker SpA} and \textit{DeepLogic Ltd}, suggesting a potential hidden relationship. Consequently, the system decides to plan a new execution, and investigate whether the two entities belong to the same corporate group. Guided by the dependency graph, the execution of the new plan (see the third block of Figure~\ref{fig:trace}) reveals that both companies are controlled by the \textit{Synthetix Global Fund}. The orchestrator therefore identifies the need for group-level aggregation, synthesizes a logical rule to aggregate the exposures and verify whether they collectively exceed the regulatory threshold (the last block of Figure~\ref{fig:trace}).
\\
\textbf{\textit{The dynamic phase highlights a concentration risk:}} while individual entities are compliant, the \textit{Synthetix Global Fund} exposes an  aggregation concentration risk, reaching $30.75\%$ of the bank’s capital.

\smallskip \noindent
\textbf{Execution and Logical Trace.}
This example illustrates a single cycle of structured and dynamic interaction, however, \vadaorchestra supports multiple iterations and plans. Such analyses are often unreliable for RAG-based or purely agentic systems, which struggle to maintain coherence as the context window becomes cluttered.
Most importantly, and in contrast to existing approaches, the system maintains a complete \textit{logical trace} in the form of a Vadalog program recording every tool invocation and synthesized rule. This trace guarantees reproducibility: evaluating it over the original data yields identical results without LLM involvement.

\section{Preliminaries}
\label{sec:preliminaries}
We first lay out some preliminary notions.

\smallskip \noindent
\textbf{Relational foundations.} A (\textit{relational}) \textit{schema} $\mathbf{S}$ is a finite set of relation symbols (or \textit{predicates}) with associated arity. A \textit{term} is either a constant or a variable. An \textit{atom} over $\mathbf{S}$ is an expression of the form $R(\bar v)$, where $R \in \mathbf{S}$ is of arity $n > 0$ and $\bar v$ is an $n$-tuple of terms. A \textit{database} (\textit{instance}) over $\mathbf{S}$ associates to each symbol in $\mathbf{S}$ a relation of the respective arity over the domain of constants and nulls.

\smallskip \noindent 
\textbf{Vadalog syntax.} Vadalog is a declarative language for ontological reasoning based on \textit{Warded Datalog$^\pm$}, a member of the Datalog family that extends Datalog with existential quantifiers while guaranteeing PTIME data complexity for query answering~\cite{gottlob2015beyond}. 
A Warded Datalog$^\pm$ program consists of a set of \textit{facts} and \textit{rules} (or \textit{tuple-generating dependencies}, TGDs). A rule has the form: $\psi(\bar x,\bar z) \mathbin{\text{:-}} \varphi(\bar x,\bar y)$, where $\varphi(\bar x,\bar y)$ (the \textit{body}) and $\psi(\bar x,\bar z)$ (the \textit{head}) are conjunctions of atoms, $\bar x, \bar y$ are universally quantified variables (quantifiers omitted), $\bar z$ is a vector of existentially quantified variables, and conjunction is denoted by comma. Vadalog extends the Warded fragment with features of practical utility. Support for aggregate functions (\textit{sum}, \textit{prod}, \textit{min}, \textit{max}, \textit{count}) is achieved via \textit{monotonic aggregations}~\cite{shkapsky2015optimizing}. Other extensions include \textit{stratified negation}, negative constraints of the form $\bot \mathbin{\text{:-}} \varphi(\bar{x},\bar{y})$ to model disjointness or non-membership, and \textit{expressions} in rule bodies with comparison $(>,<, \geq, \leq, \neq)$ and algebraic $(+,-, *, /$, etc.) operators.

\smallskip \noindent
\textbf{Reasoning and query answering.}
An \textit{ontological reasoning} task consists in answering a \textit{conjunctive query} (CQ) $Q$ over a database $D$, augmented with a set $\Sigma$ of rules. A CQ over a schema $\mathbf{S}$ has the form $q(\mathbf{x}) \mathbin{\text{:-}} \boldsymbol{\phi}(\mathbf{x}, \mathbf{y})$, where $\boldsymbol{\phi}(\mathbf{x}, \mathbf{y})$ is a conjunction of atoms and $q(\mathbf{x})$ is a predicate not in $\mathbf{S}$. A CQ $Q$ is satisfied in $D$ if there exists a homomorphism, i.e., a constant-preserving mapping $h$, from the atoms in $\boldsymbol{\phi}(\mathbf{x}, \mathbf{y})$ to the facts in $D$. The semantics of a Vadalog program is defined operationally via the \textit{chase procedure}~\cite{JohnsonK84,beeri1984proof}, which enforces the satisfaction of a set $\Sigma$ of rules over $D$ by incrementally deriving new facts until all rules are satisfied.

\section{VADAOrchestra: System Architecture}
\label{sec:system}
This section presents the technical details of \vadaorchestra. 
Figure~\ref{fig:runningexample} illustrates the overall architecture, 
highlighting the main components and their interactions.
We first describe the system architecture and the MCP 
tools. Then, we detail the 
orchestration pipeline: dependency-aware planning, plan execution, targeted data retrieval, on-demand data manipulation and replanning. Finally, we delve into the logical trace for explainability and full reproducibility.

\subsection{System Architecture}
\label{subsec:architecture}
\vadaorchestra implements a strict separation between planning and execution through a client-server architecture mediated by MCP. This design ensures that the orchestrator handles high-level planning decisions and tools parameterization delegating all data inference to the \vadalog engine.
The framework comprises two main components:
\begin{itemize}
    \item \textbf{MCP Client (Orchestrator)}: Coordinates the decision-making workflow by selecting tools, defining parameters, managing tool's dependencies, performing plan refinements, and defining data manipulations on the fly. The orchestrator serves as a coordination layer, without directly processing raw data.
    \item \textbf{MCP Server (Executor)}: Exposes reasoning capabilities as MCP tools and executes all inferences through the \vadalog engine. The server is stateless with respect to orchestration logic, it responds to individual tool invocations, and materializes results.
\end{itemize}

The communication between client and server follows the MCP specification: the client issues tool calls with typed parameters, and the server returns structured responses containing the execution status, a preview of the derived facts and some metadata. The MCP server features four categories of tools, each serving a distinct role in the pipeline.

\smallskip \noindent
\textbf{Reasoning Tools.} Reasoning tools constitute the primary interface for symbolic inference. Each reasoning tool $t_i \in \mathcal{T}_R$ is characterized by a tuple:
\begin{equation*}
t_i = \langle \textit{name}, \textit{params}, \textit{description}, \textit{pred}, \textit{schema}, \textit{deps} \rangle
\end{equation*}
where \textit{name} identifies the tool, \textit{params} specifies the required input parameters, and \textit{description} explains the purpose of the tool. Additionally, \textit{schema} defines the structure of derived facts, whereas \textit{deps} specifies the upstream tools that must be executed as prerequisites. Finally, \textit{pred} denotes the output predicate where data will be materialized. Reasoning tools execute Vadalog programs with bound parameters which encapsulate domain-specific reasoning logic (e.g., how to compute exposures from loans and credit lines).

\smallskip \noindent
\textbf{Discovery Tools.} Discovery tools enable the orchestrator to inspect the tool ecosystem and construct the dependency graph. The discovery tools are:
\begin{itemize}
    \item \textit{list\_tools()}: Returns the catalog of available reasoning tools with their descriptions, parameters, schemas, and output predicate.
    \item \textit{get\_dependencies($t_i$)}: For a given reasoning tool $t_i$, returns the set of tools $\{t_1, \ldots, t_k\}$ whose output predicates are referenced in $t_i$'s Vadalog program.
\end{itemize}

\smallskip \noindent
\textbf{Decision Support Tools.}
Decision support tools provide high-level summaries and targeted data subsets from intermediate results, ensuring the orchestrator receives only the most relevant information, thus avoiding a full data exchange between client and server that could saturate the orchestrator's context window:
\begin{itemize}
\item \textit{get\_cardinality($pred$)}: Returns the number of facts currently materialized for predicate \textit{pred}. 
\item \textit{get\_statistics($pred$,$a_i$)}: Returns statistical summaries (min, max, mean, distribution) for a specific attribute $a_i$.
\item \textit{get\_top\_k($pred$,$k$,$a_o$,$a_f$,$a_t$)}: Returns the top-$k$ facts from $pred$ ranked by attribute $a_o$, 
with optional filtering by exact match on $a_f$ and threshold 
conditions on $a_t$. This provides the LLM with a concrete, representative sample of the derived knowledge.
\end{itemize}

\smallskip \noindent
\textbf{Data Manipulation Tool.} In contrast to reasoning tools whose programs are pre-defined, the data manipulation tool accepts an arbitrary Vadalog program generated by the LLM at runtime. This allows the orchestrator to define new inference rules on-the-fly, performing joins, aggregations, and arithmetic operations over materialized predicates to address analytical 
requirements that emerge dynamically during the workflow. 
Generated rules undergo syntactic validation before execution by the \vadalog engine.

\subsection{Orchestration Pipeline}
\label{subsec:pipeline}
The orchestration pipeline unfolds through a sequence of phases that progressively construct the logical trace. The high-level control flow is illustrated in Algorithm~\ref{alg:orchestration}.

\algrenewcommand{\alglinenumber}[1]{\footnotesize#1}

\begin{algorithm}[!h]
\small
\caption{Orchestration Pipeline}
\label{alg:orchestration}
\begin{algorithmic}[1]

\Require $q$ (query), $\tau$ (cardinality\_threshold)
\Ensure $\rho$ (response), $\mathcal{L}$ (trace)
\State $\mathcal{L} \gets \emptyset, \mathcal{D} \gets \emptyset, \mathcal{D'} \gets \emptyset$ \hfill 
\State $\mathcal{T} \gets \textsc{ToolDiscovery}()$  \label{tool_discovery}
\State $\mathcal{T}_{sel} \gets \textsc{ToolSelection}(q, \mathcal{T})$ \label{tool_selection}
\State $\mathcal{I} \gets \textsc{DefineParameters}(q, \mathcal{T}_{sel})$ \label{parameterization}
\State $\Pi \gets \textsc{Plan}(\mathcal{I}, \mathcal{T})$ \label{execution_plan}

\Repeat \label{loop_start}
    \For{each stage $S_i \in \Pi$}
        \State $\mathcal{F}_{S_i} \gets \textsc{Execute}(S_i$) 
        \State $\mathcal{L} \gets \mathcal{L} \cup \textsc{Trace}(S_i$) \label{add_in_trace}
        \State $\mathcal{D'} \gets \mathcal{F}_{S_i}$
        \If{$|\mathcal{D'}| > \tau$} \hfill \label{smart_retrieval}
        \State $\mathcal{D'} \gets \textsc{GetTopK}(\mathcal{D'})$
        \State $\mathcal{D} \gets \mathcal{D} \cup \mathcal{D'}$
        \EndIf
    \State $\mathcal{M} \gets \textsc{Synthesize}(\mathcal{D}, q)$ \hfill \label{data_manip}
    \State $\mathcal{L} \gets \mathcal{L} \cup \mathcal{M}$ \label{add_in_trace2}
    \State $(\Pi, \textit{end}) \gets \textsc{Replan}(\mathcal{D}, q, \mathcal{S})$ \label{replan}
    \EndFor
\Until{$\neg \textit{end}$} \label{loop_end}

\State $\rho \gets \textsc{GenerateAns}(\mathcal{D}, \mathcal{L}, q)$ \label{answer}
\State \Return $(\rho, \mathcal{L})$
\end{algorithmic}
\end{algorithm}

\smallskip \noindent
\textbf{1. Tool Discovery and Selection.} 
The pipeline begins with the orchestrator retrieving the tool catalog $\mathcal{T}$(line~\ref{tool_discovery}) via the discovery tool \textit{list\_tools()}. Given a user query $q$, tool selection produces a subset $\mathcal{T}_{sel} \subseteq \mathcal{T}$ by prompting an LLM to identify tools whose descriptions are semantically relevant to $q$ (line~\ref{tool_selection}). Following selection, a parameterization phase extracts concrete values from the query to instantiate each tool's required parameters. For instance, given the query ``\textit{What are Bank Gamma's largest exposures?}'' and the selected tool \textit{get\_bank\_exposure(bank\_name)}, the LLM extracts ``Bank Gamma'' as the binding for \textit{bank\_name}.

Formally, a tool invocation is a pair $i = (t_j, \theta_{t_j})$ where $t_j \in \mathcal{T}$ and $\theta_{t_j}: \textit{params}(t_j) \rightarrow \Delta$ is the parameter binding, mapping each parameter to a constant from the domain $\Delta$. The set of all tool invocations, denoted by $\mathcal{I}$ (line~\ref{parameterization}), constitutes the initial step in generating a tool execution plan.

\smallskip \noindent
\textbf{2. Dependency-Aware Planning.}
Before plan execution, the system constructs a \textit{dependency graph} to identify inter-tool dependencies and ensure correct ordering. A dependency arises whenever a reasoning tool requires a predicate produced by another tool as a prerequisite.
The dependency graph $G = (\mathcal{V}, E)$ is a directed acyclic graph (DAG) where:
\begin{itemize}
    \item nodes $\mathcal{V}$ correspond to tool output predicates.
    \item edges $E$ represent dependencies: an edge $(pred_i, pred_j)$ exists if $pred_i \in \textit{deps}(pred_j)$, meaning the tool producing $pred_i$ requires $pred_j$ as input.
\end{itemize}
The graph is defined over tool output predicates rather than tool invocations, ensuring a compact dependency representation regardless of how many times each tool is invoked.

Graph construction consists in a topological sort of all the selected tools. We iterate over all selected tools  $\mathcal{I}$ and invoke \textit{get\_dependencies($t_i$)} for each. When a prerequisite tool $t' \in \mathcal{T}$ is discovered that was not in the original selection $\mathcal{T}_{sel}$, it is added to the plan and its own dependencies are resolved recursively. Since the 
tool catalog $\mathcal{T}$ is finite and $G$ is acyclic, this process is guaranteed to terminate. This mechanism guarantees that the orchestrator never omits required computations, even when the LLM's initial selection is incomplete. Tools explicitly selected by the LLM are designated \textit{high-level}; those added through dependency resolution are \textit{low-level}. The distinction propagates to their output predicates. Only high-level predicates are surfaced in the final response to preserve conciseness, while low-level predicates serve as intermediate computations retained in the logical trace for auditability.

\smallskip \noindent
\textbf{3. Execution Plan Generation.}
Given the dependency graph $G$, the system generates an execution plan (line~\ref{execution_plan}) that respects dependencies while maximizing parallelism.
An execution plan $\Pi = \langle S_0, S_1, \ldots, S_k \rangle$ is a sequence of stages, where each stage $S_i \subseteq I$ is a set of tool invocations $\langle i_0, i_1, \ldots, i_m\rangle$ that can be executed concurrently. A valid plan must satisfy the following properties:
\begin{enumerate}
    \item \textit{completeness}: $\bigcup_{i=0}^{k} S_i = \mathcal{I}$. Every tool invocation must appear in exactly one stage.
    \item \textit{disjointness}: $S_i \cap S_j = \emptyset$ for $i \neq j$. Each tool invocation is executed exactly once.
    \item \textit{dependency ordering}: for any tool invocations $(t_1,\theta_{t_1}) \in S_i$ and $(t_2,\theta_{t_2}) \in S_j$, if $(\textit{pred}(t_1), \textit{pred}(t_2)) \in E$, then $j < i$. Any tool that depends on the output of another must be scheduled in a subsequent stage.
    \item \textit{predicate-safety}: two invocations $i_1$ and $i_2$ of the same tool $t_i$ with different parameters must be placed in different stages ($S_j, S_i$ with $j \neq i$), as they write to the same output predicate.
\end{enumerate}
In our running example, the initial execution plan comprises two stages.
Stage~$S_0$ groups three independent tools $\langle$\textit{get\_bank\_info()}, \textit{get\_loans()}, \textit{get\_credit\_lines()}$\rangle$, all parameterized with ``Bank Gamma'', that execute concurrently.
Stage~$S_1$ contains $\langle$\textit{get\_bank\_exposures()}$\rangle$, which depends on \textit{get\_loans()}, \textit{get\_credit\_lines()}. Tools in~$S_0$ share no dependencies and execute concurrently;
\textit{get\_bank\_exposures()} depends on the predicates produced by
\textit{get\_loans()} and \textit{get\_credit\_lines()}, so it is
scheduled in~$S_1$.

\smallskip \noindent
\textbf{4. Iterative Plan Execution.} The execution loop iterates over the stages of the current plan until the orchestrator determines that sufficient information has been gathered (lines~\ref{loop_start}--\ref{loop_end}). Each iteration comprises stage execution, smart data retrieval, optional data manipulation, and a replanning decision.
Crucially, the orchestrator never accesses full results: it operates exclusively on targeted data samples. For each stage in the plan, the executor processes all invocations within the stage and materializes the derived facts. Whenever an invocation completes, the orchestrator receives an execution status along with some metadata (result's cardinality, output predicate, schema). This information is then used in the data retrieval stage, influencing both the subsequent data manipulation and the replanning phases.
As each stage completes, the rules corresponding to tool invocations are recorded within the logical trace~$\mathcal{L}$ (line~\ref{add_in_trace}).
For instance, after $S_0$, the trace records the first three rules of Figure~\ref{fig:trace}.

\smallskip \noindent
\textbf{5. Targeted Data Retrieval.}
A central challenge in LLM-orchestrated decision-making is scalability: tools may produce result sets with thousands or millions of facts, far exceeding what can be injected into an LLM's context window. Rather than applying fixed criteria, \vadaorchestra employs an LLM-guided retrieval strategy that adapts to the data characteristics observed at runtime. After each stage execution, the orchestrator inspects the cardinality of the materialized predicates via the \textit{get\_cardinality()} tool. Predicates whose cardinality exceeds a configurable threshold $\tau$ trigger the smart data retrieval procedure (line~\ref{smart_retrieval}), the others are directly loaded into memory instead.

\noindent\textbf{Phase 1: Statistical Analysis.}
For each large predicate, the LLM inspects the predicate schema and cardinality, along with the user's query, and decides which attributes require statistical analysis. The orchestrator then invokes \textit{get\_statistics()} for the selected attributes, obtaining distributions, ranges, and aggregates without transferring raw data.

\noindent\textbf{Phase 2: Retrieval Strategy.}
The LLM receives the computed statistics and formulates a concrete retrieval strategy: which records to retrieve, by what criteria, and how many. This strategy is expressed as a sequence of \textit{get\_top\_k()} invocations with specific ordering, filtering, and threshold parameters grounded in the observed data distribution. 

This two-phase approach results in working with representative, query-relevant data avoiding the brittleness of fixed-$k$ strategies and maintaining strict bounds on the amount of information within LLM's context window.

\smallskip \noindent
\textbf{6. On-Demand Data Manipulation.}
Pre-registered reasoning tools encapsulate domain-specific inference, but many analytical questions require combining results from multiple tools in ways that cannot be anticipated at design time. \vadaorchestra addresses this through on-demand data manipulation: the LLM generates Vadalog rules at runtime that are executed by the symbolic engine, ensuring computational correctness for joins, aggregations, and arithmetic while leveraging the LLM's ability to understand the user's intent and identify the relevant data relationships.

If the LLM determines that a manipulation is needed, it produces one or more Vadalog rules specifying joins, aggregations, or arithmetic operations over the available predicates. Since these rules reference the output predicates populated by tool invocations, they naturally operate over the complete set of derived facts
generated during the process.
The generated rules undergo syntactic validation before being dispatched to the \vadalog engine via the data manipulation tool (line~\ref{data_manip}). The results are materialized and the data manipulation rule is registered in $\mathcal{L}$ (line~\ref{add_in_trace2}).

This design achieves a clear separation of responsibilities: the LLM provides semantic understanding (what to compute), while the \vadalog engine provides computational guarantees (how to compute). Operations such as multi-way joins with arithmetic expressions and aggregations, which LLMs frequently approximate incorrectly when performed in-context, are instead executed symbolically with formal correctness guarantees.
In our example, data manipulation occurs twice.
First, the orchestrator synthesizes rules to deterministically verify whether any individual counterparty exceeds the regulatory threshold (rules 4a--4b in Figure~\ref{fig:trace}).
\noindent Since no individual exposure exceeds the limit, the
orchestrator proceeds with further investigation. After discovering that DeepLogic Ltd and NeuralLinker SpA share the same ultimate controller, it synthesises aggregation rules to assess group-level concentration risk (rules 9b--9d in Figure~\ref{fig:trace}).

\smallskip \noindent
\textbf{7. Replanning Mechanism.}
\label{subsec:replanning}
After smart retrieval and optional data manipulation, the orchestrator evaluates whether additional tool invocations are needed to adequately answer the query. The LLM receives the data collected from the previous data querying stage, the tool catalog, the invocation history, and any pending invocations, and decides whether to introduce new tool invocations (line~\ref{replan}).
When new invocations are proposed, the system performs dependency checking for the newly added tools, potentially introducing further low-level dependencies. The new invocations are integrated into the execution plan through the same predicate-safety scheduling approach used during the initial plan generation. The extended plan appends new stages to the existing plan, and execution continues from the next stage.
To ensure termination, the system limits the execution loop to a predefined maximum number of iterations and imposes a strict constraint against invoking the same tool with identical parameters twice. This prevents unbounded and unjustified replanning iterations, still providing the LLM with sufficient opportunities for data discovery and investigation.

\subsection{Logical Trace}
\label{subsec:logictrace}
A distinguishing feature of \vadaorchestra is the tracking of the reasoning process as a Vadalog program, where rules represent tool invocations and data manipulations. This trace provides a faithful, human-readable explanation of the decision-making process, and supports deterministic reproducibility without any LLM involved.
\subsubsection{Definition.}
A logical trace $\mathcal{L} = \mathcal{L}_I \cup \mathcal{L}_M$ is a Vadalog program that evolves during execution. It consists of:

\begin{itemize}
    \item \textbf{Invocation rules} ($\mathcal{L}_I$): For each needed tool invocation $i=(t_j,\theta_{t_j})$, a rule is generated whose structure directly mirrors the reasoning tool specification that we discussed in \ref{subsec:architecture}. The body contains a single atom $\textit{name}(\bar{X})$, where
    $\bar{X} = (x_1, \ldots, x_n, x_{n+1}, \ldots, x_{n+m})$:
    $x_1, \ldots, x_n$ correspond to the input parameters of the tool, as grounded by the binding $\theta_{t_j}$, and
    $x_{n+1}, \ldots, x_{n+m}$ are variables that will bind to the output actual parameters of the tool. The head is $\textit{pred}(\bar{Y})$ with
    $\bar{Y} \subseteq \bar{X}$, projecting the relevant
    subset of variables. Formally: $\textit{pred}(\bar{Y}) \leftarrow \textit{name}(\bar{X})$.
    \item \textbf{Manipulation rules} ($\mathcal{L}_\mathcal{M}$): Vadalog rules synthesized by the LLM during on-demand data manipulation, added directly to the trace after syntactic validation and execution. A manipulation rule has the form:
    \begin{equation*} h(\bar{Y},\bar{Z}) \leftarrow p_1(\bar{X}_1),\, \ldots,\, p_n(\bar{X}_n),\, C_1,\, \ldots,\, C_k. \end{equation*}
    where $h$ is a new predicate, the variables projected within $h$ are $\bar{Y} \subseteq \bigcup_i \bar{X}_i$, $\bar{Z}$ are the variables appearing in assignments in the rule body and $C_1, \ldots, C_k$ are either conditions (e.g., comparisons) over variables of the body or assignments (e.g., algebraic operations or aggregations) to variables of $\bar{Z}$. Unlike invocation rules, these rules are less constrained and may combine multiple predicates.
\end{itemize}

\subsubsection{Logical Trace Guarantees.}
The logical trace~$\mathcal{L}$ is an ordered sequence of
invocation and manipulation rules that captures the complete
execution plan: each rule records a reasoning step performed
by the system, and the sequential ordering reflects the
execution schedule determined by the orchestrator.
Since every rule in~$\mathcal{L}_I$ is generated at the moment
of tool invocation and every rule in~$\mathcal{L}_\mathcal{M}$ is
tracked after invoking the \vadalog engine, the trace faithfully records the actions the system actually performed and not a post-hoc reconstruction.
This faithfulness directly entails reproducibility:
$\mathcal{L}$ crystallizes any decision made by the LLM-orchestrator: which tools to invoke, with which parameters, in which order and which data manipulations to perform. Thereby, evaluating the trace against the same data sources allows us to re-execute exactly the same sequence of actions yielding identical derived facts without any LLM involvement, turning a sequence of LLM-based decisions into a deterministic plan. The trace is also a human-readable explanation of the entire decision-making process, providing
full auditability to domain analysts.
Figure~\ref{fig:trace} shows the complete logical trace for
the running example, illustrating the full sequence of
invocations and data manipulations that explains the
reasoning process leading to the final conclusion.

\begin{figure*}[t]
\centering
{
\fontsize{6.5}{7.5}\selectfont
\begin{tabular}{@{}c@{\hspace{2pt}}c@{\hspace{2pt}}l@{}}\toprule
\textbf{\#} & \textbf{Type} & \textbf{Rule(s)} \\
\midrule
0 & $\mathcal{L}_I$ &
  \texttt{bankInfo("Bank Gamma",TotCap,RegLim) :- get\_bank\_info("Bank Gamma",TotCap,RegLim).} \\
1 & $\mathcal{L}_I$ &
  \texttt{creditLines("Bank Gamma",Borrower,CredLim,Amount) :- 
  get\_credit\_lines("Bank Gamma",Borrower,CredLim,Amount).} \\
2 & $\mathcal{L}_I$ &
  \texttt{loans("Bank Gamma",Borrower,Amount,LoanType) :- get\_loans("Bank Gamma",Borrower,Amount,LoanType).} \\
3 & $\mathcal{L}_I$ &
  \texttt{exposure("Bank Gamma",Borrower,TotExp) :- get\_bank\_exposures("Bank Gamma",Borrower,TotExp).} \\
\midrule
4a & $\mathcal{L}_M$ &
  \texttt{regLim(BankName,RegT) :- bankInfo(BankName,TotCap,RegLim), RegT = TotCap * RegLim.} \\
4b & $\mathcal{L}_M$ &
  \texttt{exceedRegT(BankName,Borrower,TotExp) :- exposure(BankName,Borrower,TotExp), regLim(BankName,RegT), TotExp > RegT.} \\
\midrule
5 & $\mathcal{L}_I$ &
  \texttt{shareholders(Shareholder,"NeuralLinker SpA",Ownership) :- get\_shareholders(Shareholder,"NeuralLinker SpA",Ownership).} \\
6 & $\mathcal{L}_I$ &
  \texttt{ultimateCtrl(Controller,"NeuralLinker SpA") :- get\_ultimate\_controller(Controller,"NeuralLinker SpA").} \\
7 & $\mathcal{L}_I$ &
  \texttt{shareholders(Shareholder,"DeepLogic Ltd",Ownership) :- get\_shareholders(Shareholder,"DeepLogic Ltd",Ownership).} \\
8 & $\mathcal{L}_I$ &
  \texttt{ultimateCtrl(Controller,"DeepLogic Ltd") :- get\_ultimate\_controller(Controller,"DeepLogic Ltd").} \\
\midrule
9a & $\mathcal{L}_M$ &
  \textit{(rules 4a--4b repeated)} \\
9b & $\mathcal{L}_M$ &
  \texttt{ctrlExp(Controller,TotExp) :- exposure(BankName,Borrower,TotExp), ultimateCtrl(Controller,Borrower).} \\
9c & $\mathcal{L}_M$ &
  \texttt{totByCtrl(Controller,GroupExp) :- ctrlExp(Controller,TotExp), GroupExp = sum(TotExp).} \\
9d & $\mathcal{L}_M$ &
  \texttt{concRisk(Controller,GroupExp) :- totByCtrl(Controller,GroupExp), regLim("Bank Gamma",RegT), GroupExp > RegT.} \\
\bottomrule
\end{tabular}
}
\caption{Complete logical trace~$\mathcal{L}$ for the running example.
Steps 0--3: structured phase.
Step 4: first threshold check. Steps 5--8: dynamic-phase.
Steps 9: group-level concentration risk assessment.
Repeated manipulation rules are marked as such.}
\label{fig:trace}
\end{figure*}

\subsection{Answer Generation}
\label{subsec:answer}
After all execution stages are complete, the orchestrator generates the final answer through a hierarchical data injection strategy (line~\ref{answer}). Only \textit{high-level} predicates---those produced by tools selected by the LLM---are presented for answer synthesis. Low-level predicates, which served as intermediate computations and were included as dependencies of selected tools, are excluded to avoid cluttering the context. When predicates from data manipulation exist, the LLM is instructed to rely on the computed results rather than re-deriving conclusions from raw data. This ensures that the precision of the \vadalog engine's computations guides the final answer, preventing the LLM from introducing errors through in-context reasoning.

\section{Experimental Evaluation}
\label{sec:experiments}
In this section, we evaluate the performance and explainability of our framework against state-of-the-art solutions.\footnote{\vadaorchestra codebase is available upon request.}

\smallskip \noindent
\textbf{Question Answering Task.} \vadaorchestra addresses complex workflows formulated as Question Answering (QA) tasks, where a user query requires dynamic -- yet formally grounded -- iterative reasoning over a large knowledge base. We consider a QA dataset $\{q_i, a_i\}_{i\in \mathbf{N}}$ in natural language over a knowledge base $D$, where each answer $a_i$ requires multiple operations (e.g., intersections and arithmetic aggregations) to resolve the question $q_i$ (see Figure~\ref{fig:runningexample}).

\noindent\textbf{Metrics.} We evaluate the \textit{accuracy} of the framework by measuring the fraction of correctly answered questions using two complementary metrics:
(1) \textbf{Exact Match (EM)}, a case-insensitive string comparison between the predicted answer and the ground truth; and
(2) \textbf{LLM-as-a-Judge}, where an independent LLM (Llama-3.3-70B) assesses whether the predicted answer is correct and complete with respect to the ground truth, thereby avoiding penalization of semantically equivalent answers that differ only in surface form.

\smallskip \noindent
\textbf{Models and Datasets.} We evaluate our framework against four state-of-the-art agentic solutions commonly used for complex QA workflows, capturing different capabilities such as flexibility, scalability, and factual correctness: (i) \textbf{LLM}, where an LLM model answers questions using only its parametric knowledge, without access to external data;
(ii) \textbf{ReFactX}~\cite{10.1007/978-3-032-09527-5_16}, a constrained-generation approach that constructs a prefix tree over verbalized knowledge graph triples and restricts decoding to only valid fact sequences;
(iii) \textbf{LLM+RAG}, a Retrieval-Augmented Generation (RAG) system in which a retriever selects relevant data from the knowledge base and injects them into the LLM context; and (iv) \textbf{LLM+MCP}, an agentic approach where the LLM is equipped with the same reasoning tools of \vadaorchestra, exposed via MCP, but autonomously decides which tools to invoke without explicit dependency aware (re)planning or the ability to synthesize data manipulations on the fly, as seen in \vadaorchestra.
\\
Given the complexity of the workflows under evaluation, we focus on two anonymized QA datasets derived from a large European central bank knowledge graph containing approximately 10k triples and nine relations: (1) the \textit{Bank} dataset consists of 278 template-based questions (including generic, count and multi-hop QAs) introduced in \textit{ReFactX}; (2) the $\textsc{Bank}^{+}$ an augmented version for scalability stress test, where count queries require from a few up to 1,000 different KG entities. This setup enables fine-grained behavioral analysis and scalability evaluation as more relational evidence is needed to derive correct answers.
Finally, since all approaches rely on LLMs, we assess their robustness under two operational regimes: an open-weight model (Llama-3.3-70B-Instruct) and a state-of-the-art proprietary model (GPT-4o) accessed via API.

\subsection{Results and Discussion}
\begin{figure}[t]
    \centering
    \includegraphics[width=0.35\textwidth]{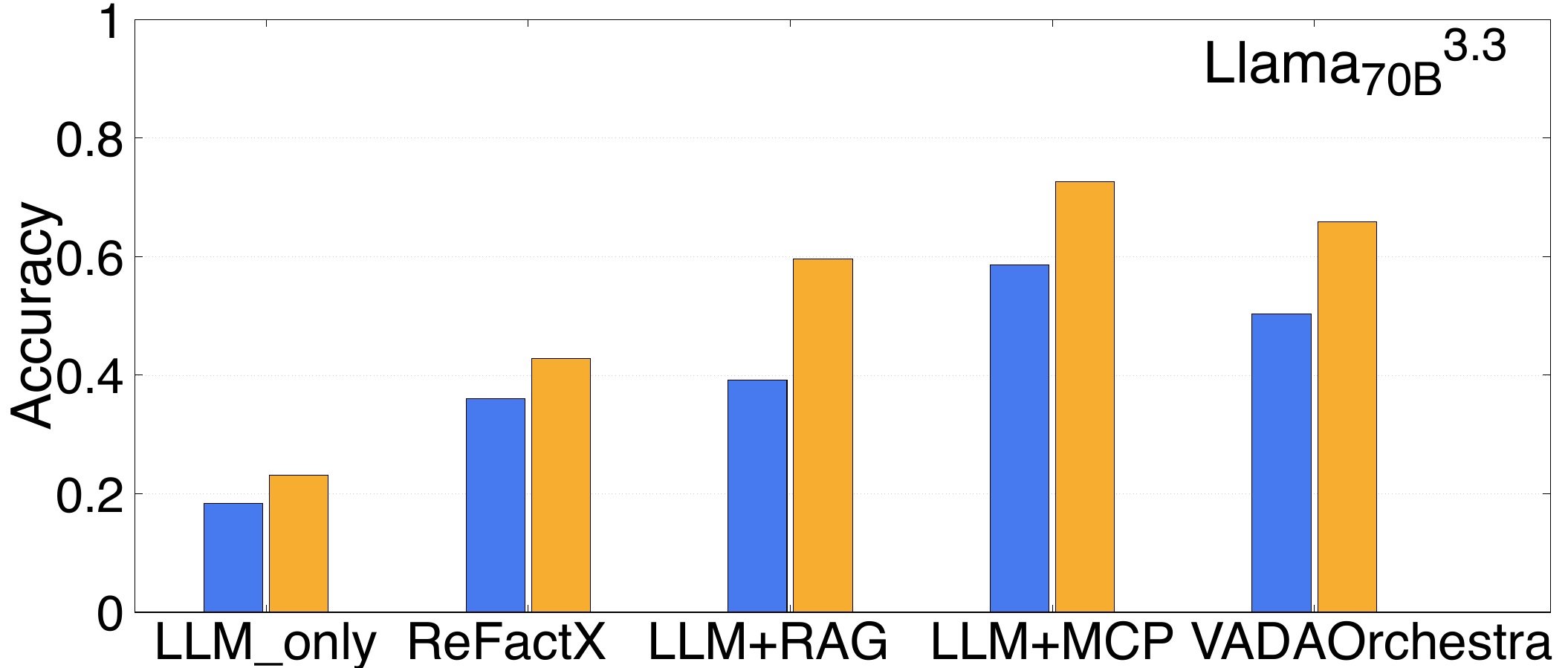}
    \includegraphics[width=0.35\textwidth]{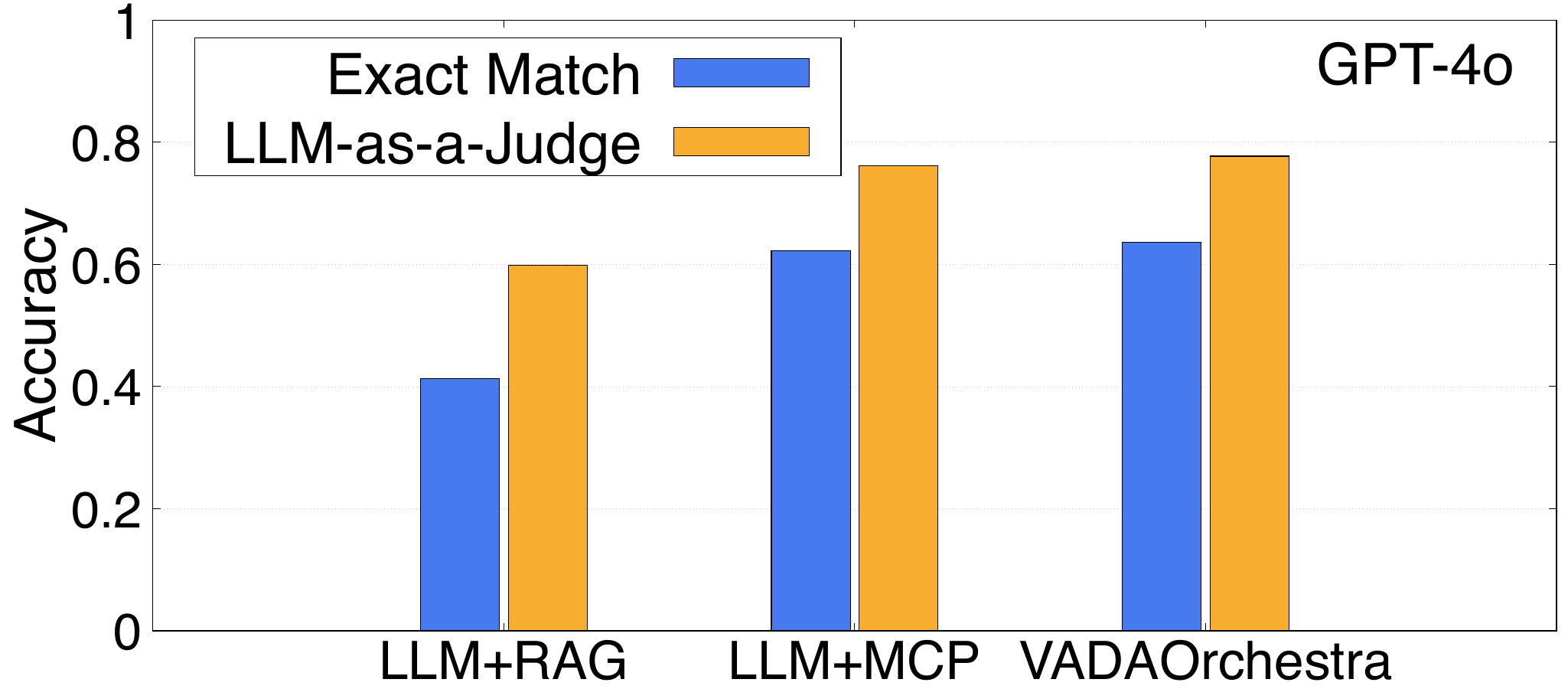}
    \caption{Results for Llama and GPT-4o across approaches.}
    \label{fig:qualitative_analysis}
\end{figure}

\textbf{Quantitative Analysis.}
Figure~\ref{fig:qualitative_analysis} reports the accuracy of all approaches on the \textit{Bank} dataset, with two different underlying LLM models. As expected, the LLM-only baseline achieves around 0.18 EM and 0.23 LLM-as-a-Judge accuracy, as the anonymized financial data lies largely outside the model’s parametric knowledge, highlighting the need for external knowledge retrieval. In fact, knowledge-enhanced approaches yield substantial performance gains. For example, \textit{ReFactX} raises accuracy to 0.36 (EM) and 0.43 (LLM-as-a-Judge), which represents a meaningful gain over the LLM-only, however, it falls short of the RAG and agentic approaches. This is because \textit{ReFactX} primarily serves to constrain the model's output to valid factual sequences, whereas \textit{LLM+RAG} and \textit{LLM+MCP} equip the model with an ``external memory'' and the agency to query it dynamically.
\textit{LLM+RAG}, which retrieves and injects relevant passages, reaches 0.39 EM and 0.60 LLM-as-a-Judge with Llama, and 0.42 EM / 0.60 LLM-as-a-Judge with 
GPT-4o---indicating that the benefit of retrieval is relatively stable across 
models.
RAG is particularly relevant in our experiments, as it represents the most established technique to mitigate context-window limitations. Nevertheless, injecting relevant passages does not address the inability of LLMs to perform reliable data-driven reasoning over large evidence sets.
Both LLM+MCP and \vadaorchestra substantially outperform traditional retrieval-based methods thanks to the available tools; however, their relative effectiveness varies with the capability of the underlying LLM model. With a Llama backbone, \textit{LLM+MCP} achieves slightly higher performance than \vadaorchestra (0.58 EM compared to 0.50), mainly due to errors in the synthesis of Vadalog rules, a core component of \vadaorchestra. Rule generation is a complex task that smaller models often handle imperfectly, producing rules that are syntactically correct but semantically flawed, thereby reducing final answer accuracy.
\begin{figure}[t]
    \centering
    \includegraphics[width=0.35\textwidth]{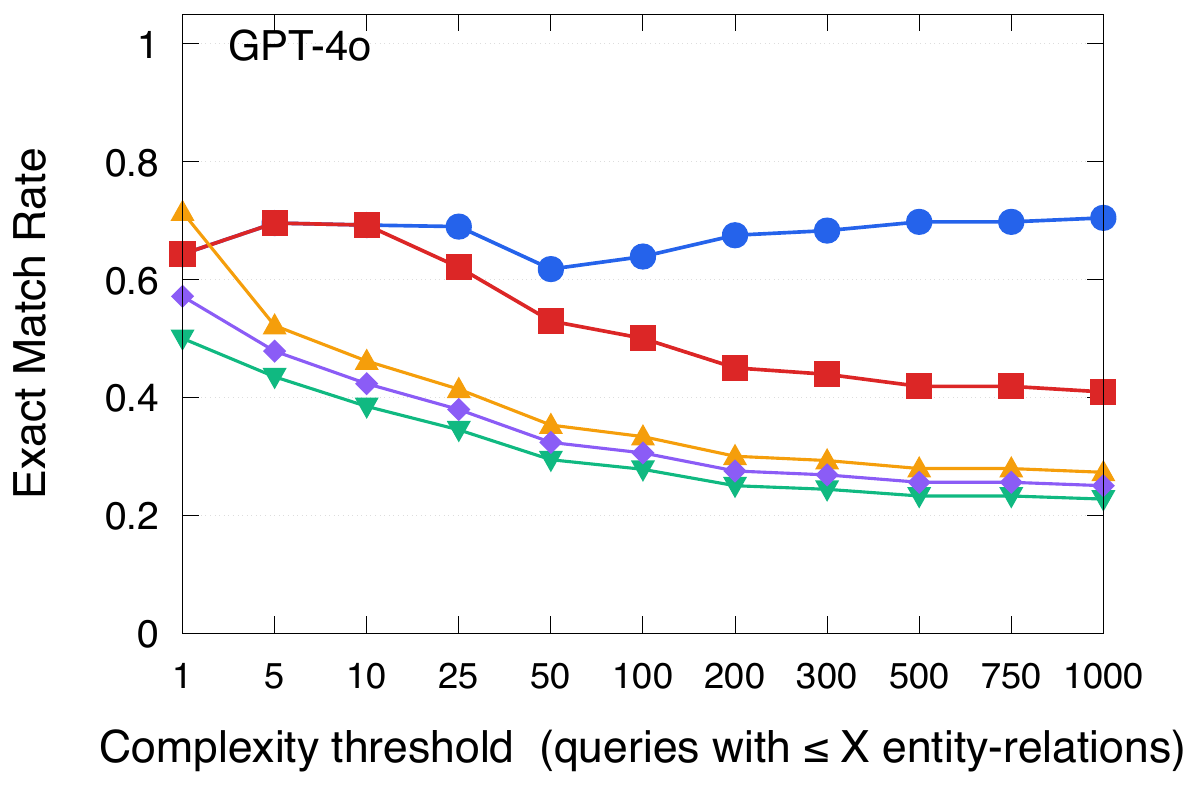}
    \includegraphics[width=0.35\textwidth]{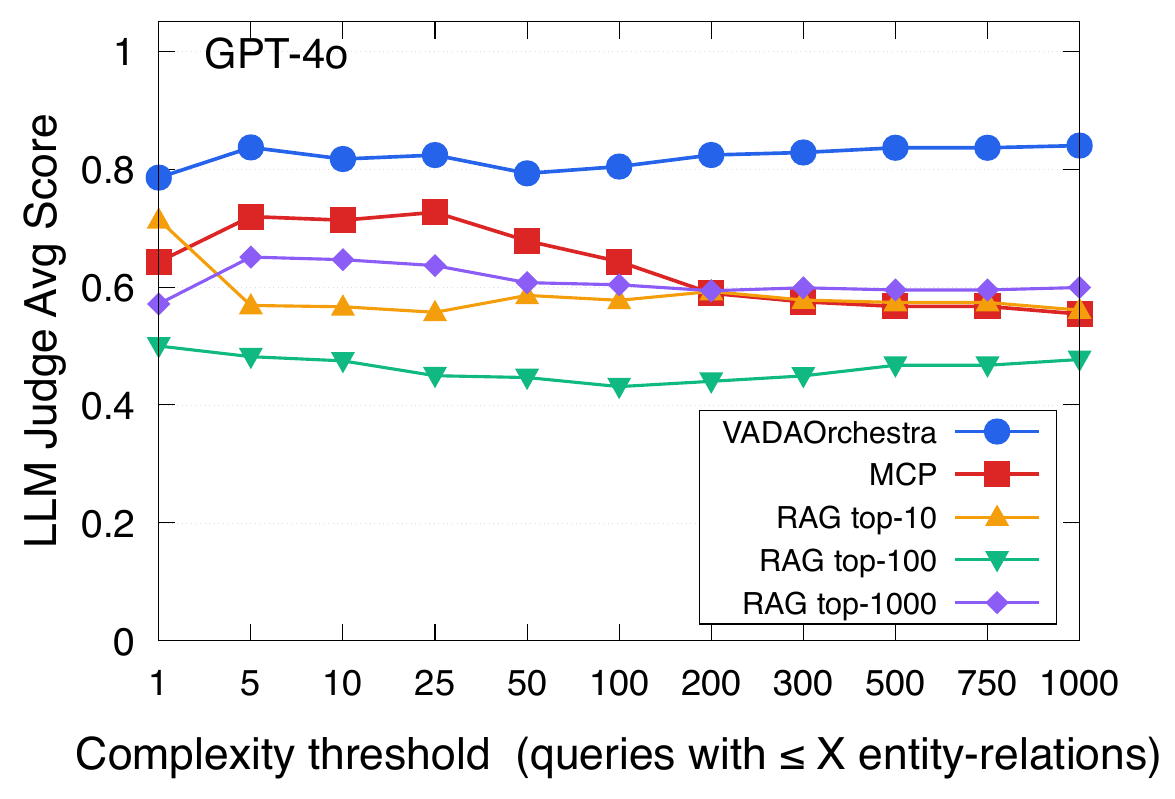}
    \caption{Cumulative accuracy plotted against query complexity.}
    \label{fig:scalability_comparison}
\end{figure}\\
However, using the more capable GPT-4o backbone, \vadaorchestra reaches 0.65 EM and 0.78 Judge accuracy, surpassing the 0.63 EM and 0.77 Judge obtained by \textit{LLM+MCP}. This indicates that \vadaorchestra better leverages stronger models to answer complex queries. Overall, the results confirm that tool-based solutions overcome single-pass retrieval and constrained-generation methods.

\smallskip \noindent
\textbf{Scalability Analysis.}
To assess how robustly each approach handles questions of increasing complexity, we evaluate the three best-performing approaches on the augmented $\textsc{Bank}^{+}$ dataset, explicitly designed to stress-test scalability, where complexity is defined by the total number of entity-relations required to compute the final answer. Figure~\ref{fig:scalability_comparison} reports the cumulative EM rate and LLM-Judge score for all queries whose complexity does not exceed a specific threshold, plotted for thresholds ranging from 1 to 1,000 entity-relations. This cumulative view shows how including progressively harder questions affects overall system performance. For \textit{LLM+RAG}, we consider three configurations, top-10, top-100, and top-1,000 records retrieved, to disentangle whether retrieval depth compensates for increasing complexity. 

In the \textbf{low-complexity regime}, the results confirm those of the quantitative analysis, with \vadaorchestra achieving the best performance, followed by the LLM+MCP and the RAG-based solutions. Notably, the LLM-as-a-Judge metric yields a score close to 0.8 for \vadaorchestra, which was slightly penalized by the EM metric.

In the \textbf{high-complexity regime}, we instead observe a clear advantage for \vadaorchestra, which maintains stable and satisfactory performance. For example, considering EM at threshold $X=50$, the RAG-based approaches performance drops to around 0.35, revealing structural limitations of retrieval constrained by the context window.
Notably, this degradation persists even when retrieval depth is progressively increased from top-10 to top-1000, confirming that enlarging the retrieved context does not compensate the limited reasoning capacity of LLMs. This is consistent with the observation that operations such as counting or verifying a property over an entire table will eventually saturate the context: while RAG can help identify the relevant records, it does not address the underlying inability of LLMs to perform such operations reliably~\cite{10.1145/3644815.3644945}.

Similarly, LLM+MCP declines significantly, from approximately 0.7 to about 0.55, demonstrating that delegating all reasoning and aggregation steps to the LLM may degrade performance in complex workflows. In most cases, these solutions produce substantially incorrect answers, drastically lowering the cumulative score despite good performance in simpler regime. By contrast, \vadaorchestra remains stable, with not less than 0.63 of EM and 0.8 LLM-as-Judge accuracy. This confirms that its scalability advantage is architectural rather than incidental, as data manipulation and logical reasoning are handled by the symbolic engine.

\begin{figure}[t]
\centering
\small
\setlength{\tabcolsep}{3pt}
\begin{tabularx}{\columnwidth}{@{}l >{\raggedright\arraybackslash}c >{\raggedright\arraybackslash}c >{\raggedright\arraybackslash}c c@{}}
\toprule
 & \textbf{Exposures} & \textbf{Controllers} & \textbf{Risk} & \textbf{Trace} \\
\midrule
\textbf{LLM+RAG} 
  & All found
  & None
  & \textcolor{red}{\ding{55}}
  & \textcolor{red}{\ding{55}} \\
\addlinespace[3pt]
\textbf{LLM+MCP} 
  & All found 
  & Identified 
  & \textcolor{orange!80!black}{$\boldsymbol{\sim}$}
  & \textcolor{red}{\ding{55}} \\
\addlinespace[3pt]
\textbf{\vadaorchestra} 
  & All found 
  & Identified
  & \textcolor{green!50!black}{\ding{51}}
  & \textcolor{green!50!black}{\ding{51}} \\
\midrule
\multicolumn{5}{@{}l}{\textbf{System conclusions:}} \\
\addlinespace[2pt]
\multicolumn{5}{@{}p{\columnwidth}@{}}{%
  \textbf{LLM+RAG}: \textit{\textcolor{red}{``Neither poses concentration risk.''}}%
} \\
\addlinespace[1pt]
\multicolumn{5}{@{}p{\columnwidth}@{}}{%
  \textbf{LLM+MCP}: \textit{\textcolor{orange!80!black}{``Ultimate controller could imply a concentration risk if exposures are considered collectively.''}}%
} \\
\addlinespace[1pt]
\multicolumn{5}{@{}p{\columnwidth}@{}}{%
  \textbf{\vadaorchestra}: \textit{\textcolor{green!50!black}{``Combined exposure \$24.6M exceeds 25\% regulatory threshold. This indicates a concentration risk due to the common ultimate controller: Synthetix Global Fund.''}}%
} \\
\bottomrule
\end{tabularx}
\caption{Running example answer comparison across approaches.}\label{tab:comparison}
\end{figure}

\smallskip \noindent
\textbf{Auditability of Answers.}
To illustrate how the approaches differ in terms of \textit{explainability}, we revisit the running example introduced in Section~\ref{sec:overview}, with Figure~\ref{tab:comparison} summarizing the answers produced by \textit{LLM+RAG}, \textit{LLM+MCP}, and \vadaorchestra. Answering the running example's question requires three main phases: (i)~retrieving all of Bank Gamma's exposures, (ii)~identifying the ultimate controllers behind each borrower, and (iii)~aggregating exposures that share a common controller to check whether their sum exceeds the 25\% regulatory threshold.

We observe that \textit{LLM+RAG} never examines the ownership structure of the borrowers and therefore incorrectly concludes that no concentration risk exists. The failure is structural: a single retrieval pass does not extend the analysis to ultimate controller relationships. \textit{LLM+MCP} correctly identifies the exposures and, through tool invocations, discovers that both \textit{DeepLogic Ltd} and \textit{NeuralLinker SpA} are ultimately controlled by \textit{Synthetix Global Fund}, suggesting potential concentration risk. However, it remains limited by its LLM-driven nature, failing to reliably perform the aggregation and threshold check required for accurate risk assessment. Finally, only \vadaorchestra produces the correct answer. After identifying all exposures and discovering the ultimate controllers via dedicated tool invocations, it delegates the aggregation and threshold check to the \vadalog engine: the total exposure of \$24.6M under \textit{Synthetix Global Fund} is computed deterministically and flagged as exceeding the regulatory limit.
\\
Beyond correctness, the distinguishing feature of \vadaorchestra is the logical trace it produces, shown in Figure~\ref{fig:trace}.
The analyst can therefore interpret the final answer back through each intermediate step, a property that is indispensable in regulated financial settings.

\smallskip \noindent
\textbf{Limitations in Real-World Financial Use Cases.}
We evaluated \vadaorchestra across a wide range of financial use cases---including 
company ownership and concentration risk assessment---without 
encountering major shortcomings in terms of accuracy. We nonetheless identified two 
main limitations that are worth discussing.
First, when the system processes entities that, at runtime, turn out to have many 
relationships relevant to the question, we observe an increase in token 
generation due to the multiple LLM invocations. Consequently, the overall execution time 
of \vadaorchestra grows with respect to a traditional agentic approach, since rule 
synthesis introduces further overhead. We consider this an acceptable trade-off in light of the gains in correctness and 
scalability.
Second, the quality of the generated data manipulation rules can degrade in complex financial scenarios, when smaller language models 
are used as the backbone (see Figure~\ref{fig:qualitative_analysis}).
In future work, we plan to extend the system with additional validation mechanisms to improve the quality of the generated rules, thereby enhancing the performance of small language models.

\section{Conclusion}
\label{sec:conclusions}
This paper introduces \vadaorchestra, a novel neurosymbolic framework for automating knowledge-intensive workflows that require both adherence to structured domain procedures and flexibility to adapt to runtime findings. In particular, \vadaorchestra combines an LLM-driven orchestration with symbolic reasoning: every tool invocation and data manipulation is encoded as a logical rule, representing the entire decision-making process as an evolving logical program that provides a faithful, explainable, and reproducible execution trace.
Experimental results on a real-world financial dataset demonstrate strong accuracy and auditability compared to purely agentic baselines. Future work will focus on integrating validation mechanisms to enhance the quality of synthesized rules, with a specific emphasis on improving the performance of small language models.

\newpage

\section*{Acknowledgments}
This research was kindly supported in whole or in part by the Vienna Science and Technology Fund (WWTF), grant numbers: 10.47379\slash VRG18013, 10.47379\slash ICT25032, 10.47379\slash NXT22018, 10.47379\slash ICT2201, 10.47379\slash DCDH001 as well as by the Austrian Science Fund (FWF) under grant number 10.55776\slash COE12.
\section*{AI Declaration}
The authors have not employed any Generative AI tools.

\bibliographystyle{kr}
\bibliography{biblio}

\clearpage
\clearpage

\appendix

\end{document}